# Transformationally Identical and Invariant Convolutional Neural Networks through Symmetric Element Operators

*ShihChung B. Lo, Ph.D.[1,2,3], Matthew T. Freedman, M.D.[2,3], Seong K. Mun, Ph.D.[2], and Shuo Gu, Ph.D.[4]

[1] Radiology Department, Georgetown University Medical Center, Washington DC 20007
[2] Arlington Innovation Center: Health Research, Virginia Tech, Arlington VA 22203
[3] Oncology Department, Georgetown University Medical Center, Washington DC 20007
[4] HIT Alibaba Cloud Data Science Academy, Harbin Institute of Technology, Huizhou, Guangdong, China

*e-mail: <dcben0@gmail.com> or <benlo@vt.edu>

## Abstract

Mathematically speaking, a transformationally invariant operator, such as a transformationally identical (TI) matrix kernel (i.e., K = T{K}), commutes with the transformation (T{}) itself when they operate on the first operand matrix. We found that by consistently applying the same type of TI kernels in a convolutional neural networks (CNN) system, the commutative property holds throughout all layers of convolution processes with and without involving an activation function and/or a 1D convolution across channels within a layer. We further found that any CNN possessing the same TI kernel property for all convolution layers followed by a flatten layer with weight sharing among their transformation corresponding elements would output the same result for all transformation versions of the original input vector. In short, CNN[ Vi ] = CNN[ T{Vi} ] providing every K = T{K} in CNN; where Vi denotes input vector and CNN[·] represents the whole CNN process as a function of input vector that produces an output vector. With such a transformationally identical CNN (TI-CNN) system, any transformation, that is not associated with a predefined TI used in data augmentation, would inherently include all of its corresponding transformation versions of the input vector for the training. Hence the use of same TI property for every kernel in the CNN would serve as an orientation or a translation independent training guide in conjunction with the error-backpropagation during the training. This TI kernel property is desirable for applications requiring a highly consistent output result from corresponding transformation versions of an input. Several C programming routines are provided to facilitate interested parties of using the TI-CNN technique which is expected to produce a better generalization performance than its ordinary CNN counterpart.

**1. Introduction -**

The initial structures and algorithms of convolutional neural networks (CNN) were motivated by neocognitron (Fukushima 1980) and error-backpropagation (Rumelhart 1986) and were independently developed by three groups in late 1980s and early 1990s (Zhang 1988; 1990; LeCun 1989; 1998; and Lo 1993; 1995(a); 1995(b)). The convolution neural network was first named by Lo et al (1993) who consistently developed and applied CNN related techniques to medical imaging pattern recognition and analysis research such as detection of lung nodule/cancer and breast lesion/cancer. Later, the same group developed two other CNN systems: the CNN with wavelet kernels (Lo 1995(c)) and the CNN with circular kernels (Lo 1998; 2002). Rotation and transpose of input vectors as a part of data augmentation were introduced in their early papers as well (1993; 1995(a); 1995(b)). The CNN with symmetrical kernels was first presented by the same authors for recognition of circular and radial correlated vectors such as breast lesions on mammogram (1998; 2002). The CNN for image segmentation was introduced by Hasegawa and Lo et al (1994; 1998). Having studied a variety of convolution kernels, a unified theoretical framework through the dyadic decomposition was developed for filter banks by Lo and his co-workers (1997).

Since then modifications of CNN structure were proposed by many investigators, particularly in recent CNN research and applications. Among them, methods to achieve transformationally invariant, mainly in rotation and/or translation, have been attempted: Gens and Domingos (2014); Dieleman et al. (2015; 2016); Cohen et al. (2016); Cheng et al (2016); Ravanbakhsh et al. (2017); Zaheer et al. (2017b); Guttenberg et al. (2016); Cohen et al. (2017); Cohen et al. (2018). Rotation invariant based on Fourier Transform for a CNN was proposed by Marcos et al. (2016) and Chidester et al (2018).

This paper is intended to exploit families of transformationally identical (TI) vectors and their mathematical foundations that can make CNN produce a quantitatively identical result through a series of processes in the CNN when a transformation of the input vector does not involve interpolation (type-1 transformation). In the same time, transformations of the input vector involving interpolation (type-2 transformation) would be used to train the CNN to produce qualitatively invariant output results. This is because the latter alters the original input vector to the begin with, therefore it would be technically incapable of maintaining computationally identical results within a CNN process. In this paper, our strategies for constructing the CNN and making it to produce a highly consistent performance are:

a) type-1 transformations of an input vector would inherently include in the TI-CNN process;
b) the use of type-2 transformations of the input vector and their TI counterparts is intended to produce qualitatively invariant output results through the TI-CNN training.

Note that the use of each type-2 transformation would inherently include its associated type-1 transformation counterparts for training which is significantly different from conventional data argumentation approach. With the new method, each transformation automatically works as a group in the CNN. However, each of conventional transformation works on its own.

This mathematical property of TI is not limited to the convolution process but also observed in many functions with scan operation operating on the first operand space which will be reviewed in section 2.1 first followed by convolution computation with symmetric kernels and associated CNN processes.

## 2. Methods -
2.1. The Property of Transformationally Identical Vector – An unified theory of symmetric operator

Considering a vector ($V_i$) operated by an operation " ! " with a kernel K scanning through the entire $V_i$ space that results in another vector $V_{r1}$, as

$$V_{r1}(x',y',z') = V_i(x,y,z) \,!\, K(u,v,s). \qquad ...(1)$$

where (u,v,s) denotes each element of K containing either a value or a function. Mathematically there are interesting transformation characteristics worth exploring.

With a specific geometry transformation characteristics of K, the resultant vector ($V_{r1}$) would possess the same transformation property that can apply to vector ($V_i$) to share the computation without recomputing its transformation counterparts T{$V_i$}, by providing

$$K(u_t,v_t,s_t) = T\{\, K(u_t,v_t,s_t) \,\} \qquad ...(2)$$

Let T**{.}** be a vector transformation, RT**{.}** is its corresponding fully reversible transformation, we have.

$$V_i(x,y,z) = RT\{ T \{ V_i(x,y,z) \} \} \qquad \ldots(3)$$

The resultant vector can be expressed as

$$V_{r1}(x',y',z') = RT\{ T\{ V_{r1}(x',y',z') \} \}$$

$$= RT\{ T\{ V_i(x,y,z) \;!\; K(ut,vt,st) \} \}$$

$$= RT\{ T\{ V_i(x,y,z) \} \;!\; T\{ K(ut,vt,st) \} \} \qquad \ldots(4)$$

By replacing eq. (4) with (3)

$$V_{r1}(x',y',z') = RT\{ T\{V_i(x,y,z) \} \;!\; K(ut,vt,st) \}. \qquad \ldots(5)$$

Because of eq. (1), we have

$$V_i(x,y,z) \;!\; K(ut,vt,st) = RT\{ T\{ V_i(x,y,z) \} \;!\; K(ut,vt,st) \}. \qquad \ldots(6)$$

or $T\{ V_i(x,y,z) \;!\; K(ut,vt,st) \} = T\{ RT\{ T\{ V_i(x,y,z)\} \;!\; K(ut,vt,st) \} \}$

which is equivalent to

$$T\{ V_{r1}(x',y',z') \} = T\{ V_i(x,y,z) \;!\; K(ut,vt,st) \}$$

$$= T\{ V_i(x,y,z) \} \;!\; K(ut,vt,st) \qquad \ldots(7)$$

Equation (7) indicates that a specific transformation of resultant vector $V_{r1}$ obtained by $V_i \;!\; K$ which is equal to the transformation of the input vector ($V_i$) (i.e., $T\{ V_i \}$) operates with the same kernel (K). However, the kernel must possess the property of $K = T\{ K \}$.

To expand this property, the operation can be generalized to represent a set of operations (S) followed by the scan operation as

$$s(x',y',z') = S[ V_{ri}(x',y',z') ] \qquad \ldots(8)$$

that is geometry transformationally independent. In other words, S and T operations commute.

Hence equation (4) can be expressed as

$$T\{ s(x',y',z') \} = T\{ S[ V_{r1}(x',y',z') ] \}$$

$$= S[ T\{ V_i(x,y,z) \;!\; K(ut,vt,st) \} ]$$

$$= S[ T\{ V_i(x,y,z) \} \;!\; K(ut,vt,st) ] \qquad \ldots(9)$$

Similar to eq. (7), equation (9) indicates that a specific transformation of resultant vector "s" obtained by $V_i \;!\; K$ followed by the S operation which is equal to the transformation of the input vector $V_i$ (i.e., $T\{ V_i \}$) operates with the same kernel "K" followed by a set of operations "S". Again, the K must possess the property of $K = T\{ K \}$. We can call this property as transformationally invariant quantitatively, or transformationally identical (TI). This quantitatively identical property through a predefined transformation and it applications to CNN is the main focus of this paper.

Both eqs. (7) and (9) imply that given a geometrical transformation, using any K(ut,vt,st) as operation kernel that is quantitatively identical to its transformation counterparts, we can expect an identical result by entering the same transformation of the input vector (Vi) into a chain of processes involving scan operation and transformationally independent operations.

In the field of image processing, there are several commonly used scan operations: convolution, correlation, wavelet decomposition, adaptive histogram equalization, morphological operation (erosion, dilation, open, closing, morphological gradient, etc), sampling, and pooling for neural network processing. Equation (9) can be held for all these scan operations, should the requirement of K = T{ K } be satisfied. Hence an unified theory of transformationally identical processing can be held if the same family of symmetric operators are employed throughout the entire operation pipeline.

## 2.2. Applications of Transformationally Identical (TI) Property to Digital Signal Processing

Two main categories of geometry transformation can be made so that a kernel "K" can be found and satisfy the requirement condition of K= T{ K }.

1) Linear translation of a given direction:
Linear translations without interpolation for a 3D vector (type-1 transformation).

1.1) Based on each of x=0, y=0 and z=0 plains
(a) along $0°$, (b) along $45°$, (c) along $90°$, and (d) along $135°$.

1.2) Along each of 4 lines:
(a) along x=y=z, (b) along x=-y=z, (c) along x=-y=-z, and (d) along x =y=-z

2) Rotation and Reflection Symmetry In Cubic Matrix
Transformations without interpolation for a 3D vector (type-1 transformation).

2.1) Based on each of x=0, y=0 and z=0 plains

(a) $90°$ rotation ($R_i$<-> $R_{i+1}$, where i = 0, 1, 2)

(b) $180°$ rotation or 2D centrosymmetry ($R_i$ <-> $R_{i+2}$, where i = 0, 1)

(c) left right reflection symmetry (m1)

(d) top down reflection symmetry (m2)

(e) upper-right lower-left reflection symmetry (d1)

(f) upper-left lower-right reflection symmetry (d2)

Composed: 2D dihedral symmetry of order 8 (Dih4 or D8 - composed of reflection and $90°$ rotation) which also covers the above 7 types of transformations

2.2) Based on each of 6 axes: (x=0, y=z); (x=0, y=-z); (y=0, z=x); (y=0, z=-x); (z=0, x=y); (z =0, x=-y);
(a) $180°$ rotation

2.3) Based on each of 4 axes: x=y=z, x=-y=z, x=-y=-z, and x =y=-z
(a) $120°$ rotation
(b) $240°$ rotation

2.4) Based on x=y=z=0

(a) centrosymmetry

Composed: 3D dihedral symmetry of order 8 (Dih4 or D8 - composed of reflection and 90$^o$ rotation) which also covers the above 33 types of transformations.

2.5) Dih4 symmetry with biorthogonal wavelet kernels

In a Dih4 TI kernel, element coefficients at a corner wedge are free parameters. Elements on other wedges corresponding to the Dih4 symmetric element positions in the wedge would share the same value. Though wavelet decomposition is different from an ordinary convolution process, the biorthogonal kernels for each compartment though may be different but the absolute value in each element of kernel are the same. In addition, the multi-dimensional wavelet decomposition is made by one-dimensional convolution process and down sampling 1/2 at a time. The total number of free parameters is much less (the number of elements in 1D kernel plus 1 divided by 2). In effect, kernels to produce low-low (LL) and high-high (HH) compartments are Dih4 TI kernels. Kernels to produce low-high (LH) and high-low (HL) compartments are Dih4 TI with an odd number of elements but are anti-symmetric (i.e., 180$^o$ rotation TI) with an even number of elements. Since each compartment is processed through an independent pipeline in the neural network process; for latter situation, there is still room to make signals from LH and HL be Dih4 TI, if desired. This can be done by inserting a refection TI kernel in each of these two compartment pipelines.

2.3. Applications of TI to the computation of Convolution Neural Networks (CNN) - Transformationally Identical CNN (TI-CNN)

2.3.1. Convolution layer with an activation, stride, pooling layer and filtering across the channels in a typical CNN

Because the convolution process in convolution layers of a CNN belong to a kernel scan operation over input vector $V_i(x,y,z)$, equation (4) holds as long as K = T**{** K **}**.

T**{** $V_{r1}(x',y',z')$ **}** = T**{** $V_i(x,y,z)$ * K(ut,vt,st) **}**

= T**{** $V_i(x,y,z)$ **}** * K(ut,vt,st)  ...(10)

where " * " denotes as a convolution operation. Figure 1 shows two convolution processes: (i) with a kernel that possesses no symmetry (Kp); (ii) with a kernel that is Dih4 symmetric (Kq). Considering activation function that is applied to each node of convolution channel:

a(x',y',z') = A**[**$V_{r1}(x',y',z')$**]**  ...(11)

Since the activation function, stride, and pooling are transformationally independent, we have

T**{** $s_1(x',y',z')$ **}** = T**{** C**[** $V_{r1}(x',y',z')$ **]** **}**

= C**[** T**{** $V_i(x,y,z)$ **}** * K(ut,vt,st) **]**  ...(12)

where C[·] is the composed function of activation function A[·], stride, 1-D filtering across channels in the same layer (also called 1x1 convolution), and pooling operation for each element on the resultant array (x',y',z'). However, the stride distance and pooling operation size must be able to evenly divide the size of receptive vector space in all dimensions to make sure operations are performed at all corresponding transformationally symmetric positions systematically.

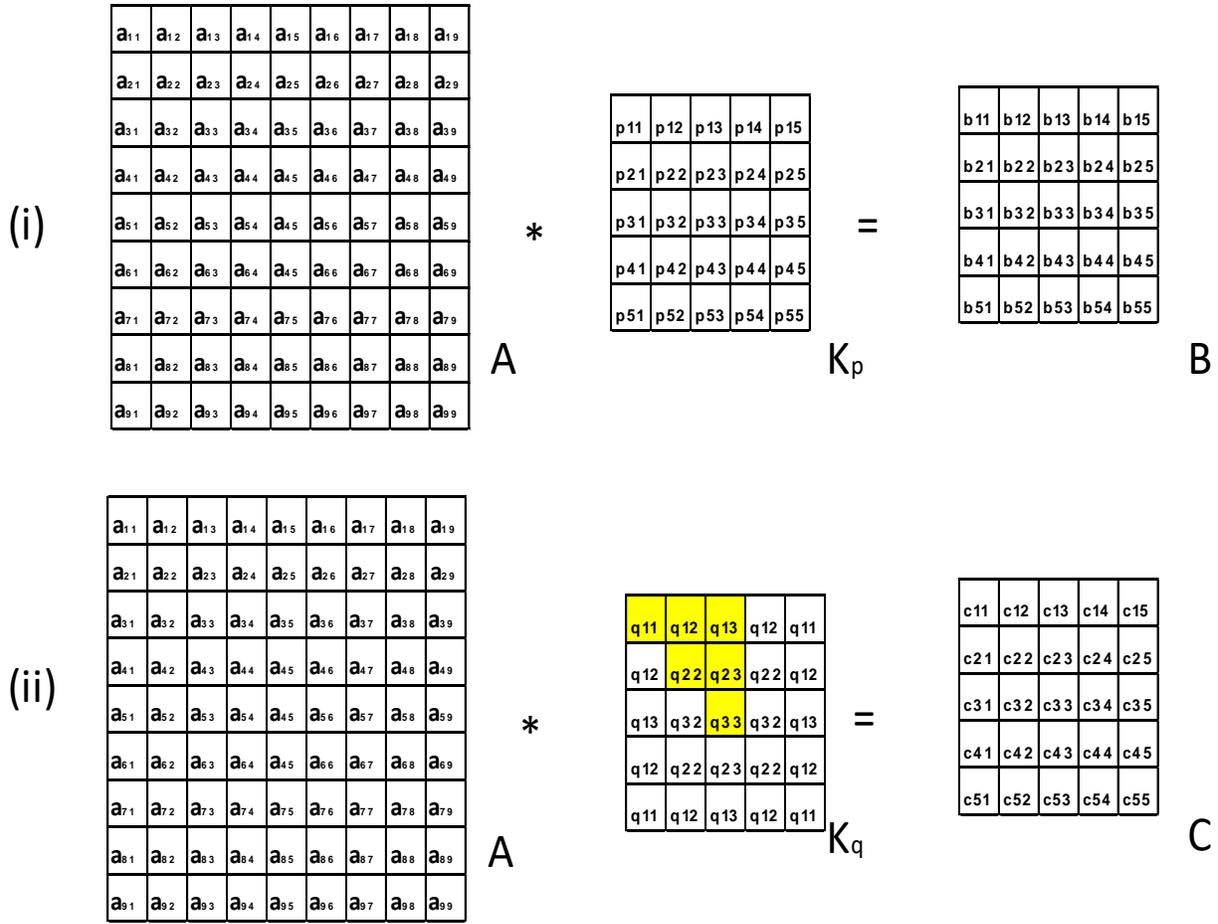

Figure 1. Two convolution processes: (i) with a kernel that possesses no symmetric property (Kp); (ii) with a kernel that is Dih4 symmetric (Kq). When matrix A rotates with any of Dih4 symmetric group, the resultant matrices B and C have different transformation characteristics due to the difference characteristics of their corresponding kernels (i.e., Kp and Kq), respectively:

(i) $A * K_p = B$ but $T_{Dih4}\{A\} * K_p \neq T_{Dih4}\{B\}$ because $K_p \neq T_{Dih4}\{K_p\}$.
(ii) $A * K_q = C$ and $T_{Dih4}\{A\} * K_q = T_{Dih4}\{C\}$ because $K_q = T_{Dih4}\{K_q\}$.

Note that elements marked in yellow are free parameters in Kq, the rest of elements grouped in other 7 wedges are Dih4 symmetric to the wedge marked in yellow. In 3D, there are 24 cone wedges with the Dih4 symmetry (8 versions of rotation/reflection on each plain).

2.3.2. Introduction of sharing weight for all corresponding transformational positions on the flatten layer

Each node (n) in the fatten layer is a channel (condensed feature), the resultant vector is a single value. Therefore, its operation, usually does not involve stride and pooling, is a special case of equation (12). Again providing $K_f$ possesses a transformationally identical property.

$$K_f(n; ut, vt, st) = T\{ K_f(n; ut, vt, st) \} \qquad ...(13)$$

we have

$$T\{ s_2(n; x'', y'', z'') \} = T\{ A[ V_{r2}(x'', y'', z'') ] \}$$

$$= A[\ T\{\ V_{r1}(x',y',z')\ \}\ *\ K_f(n;\ ut,vt,st)\ ] \quad ...(14)$$

Since either $s_2$ or $V_{r2}$ only contains one single value, the transformation is the same as the identity function. In addition, the dimension of $K_f$ is the same as $s_1$ and $V_{r1}$ (i.e., (x',y',z')), we have

$$s_2(n) = A[\ V_{r2}\ ] = A[\ T\{\ V_{r1}(x',y',z')\ \}\ \bullet\ K_f(n;\ x',y',z')\ ] \quad ...(15)$$

where $V_{r2} = s_1(x',y',z')\ \bullet\ K_f(n; x',y',z')$

or $V_{r2} = C[\ V_{r1}(x',y',z')\ ]\ \bullet\ K_f(n; x',y',z')$

Since $K_f$ shares weights on all corresponding transformational positions, when it times (inner product) a vector of same size, it possesses an interesting property that would produce the same value with or without performing the transformation of the first operand as

$$T\{\ s_1(x',y',z')\ \}\ \bullet\ K_f(n;\ x',y',z') = s_1(x',y',z')\ \bullet\ K_f(n;\ x',y',z') \quad ...(16)$$

Replacing (13) into equation (12) we have

$$s_2(n) = A[\ T\{\ s_1(x',y',z')\ \}\ \bullet\ K_f(n;\ x',y',z')\ ] = A[\ s_1(x',y',z')\ \bullet\ K_f(n;\ x',y',z')\ ] \quad ...(17)$$

Replacing (7) to (14), we have

$$s_2(n) = A[\ C[\ T\{\ V_i(x,y,z)\ \}\ *\ K(ut,vt,st)\ ]\ \bullet\ K_f(n;\ x',y',z')\ ] \quad ...(18)$$

Equation (18) indicates that the TI property is fully preserved without and with a predefined transformation for the input vector ($V_i$).

2.3.3. CNN as a composed function

A typical CNN takes a vector from the input layer, processes it with convolutional processes etc, and produces a resultant vector on the output layer. On the first flatten layer which uses shared weights corresponding to their transformation element positions and produces inner product between weights and channel nodes on the last convolution/pooling layer, the fully connected nodes in the further layer would perform as a regular multi-layer perceptron (MLP) processing and finally reach to the output layer. Having a CNN trained by this manner, the output of the CNN is identical with the same type of TI kernels for all corresponding transformations of $V_i$. Hence it is a transformationally identical CNN (TI-CNN) when it possesses a set of TI kernels $K_s = T\{\ K_s\ \}$ on all convolution layers and weights sharing on the first flatten layer in the CNN pipeline. We can call such a CNN a TI network system (or function)

$$TI\text{-}CNN[\ V_i\ ] = TI\text{-}CNN[\ T\{\ V_i\ \}\ ] = V_o \quad ...(19)$$

as long as every kernels operated inside the CNN possesses the same transformation or its effective composition property ($K = T\{\ K\ \}$). Therefore, there is no need to use predefined corresponding transformations of the input vector to enter into the input layer for such a TI-CNN system. Because they all end-up the same result at each node on and after the first flatten layer.

The advantage of using this kind of TI-CNN has 4-fold:
1) The use of a specific TI property must be based on characteristics of the input and output vectors bearing a training task for the categorization or analysis of input data distribution. Besides error-backpropagation, the coefficients of each kernel with a TI property in the whole

CNN are trained in a highly coordinated manner rather than training of unconstrained coefficients in a conventional CNN. It is expected that these TI convolution kernels may be able to extract features associate with the TI likely representing a symmetrical property in the input and reflecting on the output.

2) No need to use inclusive transformation versions of input vector T**{** Vi **}** as a part of data augmentation for the TI-CNN training.

3) The total effective number of elements in a kernel are reduced by a factor of 3-6 when small kernels are used. Hence the interpretation of filter characteristics is easier than unsymmetrical ones. This may help explain feature learning in a CNN.

4) During the training, the TI-CNN intrinsically takes multiple input versions with each type-2 transformation input. For example, by rotating the Vi $22.5^o$ and entering it onto the input layer, the training for a 2D CNN with Dih4 kernels would automatically gain a total of 8 versions including $22.5^o$, $112.5^o$, $202.5^o$, $292.5^o$ and their reflection. If we rotate Vi $10^o$, $20^o$, $30^o$, $40^o$, $50^o$, $60^o$, $70^o$, $80^o$ and their reflection for data augmentation, the TI-CNN would receive a total of 72 Vi versions of training including each $10^o$ increment version and its corresponding Dih4 symmetry versions. The use of type-2 transformation of the input vector is intended to achieve qualitatively invariant output results of the CNN system consistently as used in an ordinary CNN training. However, when both type-1 and type-2 transformations are used in the proposed TI-CNN system, the TI-CNN is expected to produce greater performance in generalization.

2.3.4 Inclusive of kernel types for maintaining the same TI property.
      For any rotation based transformation with a given type of RTI kernels, Dih4 TI kernels can always support the RTI system. The RTI-CNN as a whole would maintain the same TI system. For a $180^o$ rotation TI system, both $180^o$ rotation and $90^o$ rotation TI kernels can be used and the CNN as a whole would maintain a $180^o$ rotation TI system. In addition, the TI-CNN can also be generalized so that each kernel process can operate with a predefined operation as discussed in section 2.1 rather than an ordinary convolution operation.

2.3.5 Translationally identical (TLI-CNN) property in convolution layers and the first flatten layer
      Without considering stride and pooling, type-1 translationally identical property maintains on intermediated resultant matrices in all convolution process intrinsically without any constraint to the operation kernel. Therefore, the simplest way to achieve translationally identical CNN is to arrange the weight sharing for fan-in connection at each node of the first flatten layer. There are 4 type-1 translationally identical groups in a 2D CNN. The fan-in weight sharing through full network connections can be arranged parallelly to the translation direction.

1) Horizontal translation (y=0)
      $s_2(n)_{y=0}$ = A**[** C**[** T**{** Vi(x,y) **}** * K(u,v) **]** · Kf(n; x', y'=0) **]**       ...(20)
      where Kf(n; x', y'=0) share the same weights along y=0

2) Vertical translation (x=0)
      $s_2(n)_{x=0}$ = A**[** C**[** T**{** Vi(x,y) **}** * K(u,v) **]** · Kf(n; x'=0, y') **]**       ...(21)
      where Kf(n; x'=0, y') share the same weights along x=0

3) Along $45^o$ translation (x=y)
      $s_2(n)_{y=x}$ = A**[** C**[** T**{** Vi(x,y) **}** * K(u,v) **]** · Kf(n; x'=y') **]**       ...(22)
      where Kf(n; x'= y') share the same weights along x=y

4) Along 135° translation (x=-y)

$$s_2(n)_{x=-y} = A\mathbf{[}\ C\mathbf{[}\ T\mathbf{\{}\ V_i(x,y)\ \mathbf{\}} * K(u,v)\ \mathbf{]} \cdot K_f(n; x'=-y')\ \mathbf{]} \quad ...(23)$$

where $K_f(n; x'=-y')$ share the same weights along x=-y

It is worthwhile to mention that all Kf's described above must have the same size as the last resultant matrix before the first flatten layer. The major difference between these 4 translationally identical groups and the 2D version of eq. (18) is that the convolution kernel K(u,v) does not require any constraint (i.e., all elements of K(u,v) are free parameters). As indicated in section 2.1, there are other type-1 translation groups without altering digital values of the input vector in 3-D space. They all have a similar translationally identical process corresponding to eqs. (20) - (23) when a 3D CNN is employed.

For the implementation of a TLI-CNN, the minimum padding region (filling extended array elements with the same value) along the translation direction is the intended maximum translation distance times the original input vector size for 0° and 90° direction TLI-CNNs; and is the extended frame region of the original matrix associated with maximum translation distance for 45° and 135° direction TLI-CNNs. With such a padding, all scanning processes operating the first operand such as convolution in the CNN would cover the same receptive area on the original vector though being translated.

However, only when those translation distances that are evenly divisible by the effective length of stride times pooling size, would the TLI property holds in the CNN, otherwise the TLI property could not be held in the CNN. This is because stride and/or pooling would not operate on the same receptive area unless translation distance of the input vector is the same distance or its even order.

2.3.6. Wavelet decomposition as a part of convolutional process

In 1995, Lo et al has introduced a CNN structure using wavelet kernels and their associated decomposition compartments. The biorthogonal wavelet family possesses some interesting symmetric properties as far as TI is concerned. Typically, the scaling filter of a biorthogonal wavelets is symmetric. However, its wavelet filter may be symmetric or anti-symmetric. For a symmetric wavelet filtering, Dih4 type TI property can be observed in LL, LH, HL, and HH decomposition compartments. With a symmetric wavelet filter (i.e., odd number of kernel elements), Dih4 TI filters can be used and its corresponding transformation property can be observed in all 4 decomposition compartments. However, only 180° rotationally identical symmetry would be observed in LH and HL decomposition compartments if an even number of kernel elements is employed. Therefore, the whole CNN processed by this type of biorthogonal wavelet kernels can be made with 180° rotationally identical. By inserting an additional reflection TI kernel in each of LH and HL compartment pipelines, the resultant channel would turn into a Dih4 TI property system, if desired.

After the wavelet kernel decomposition, a conventional convolutional neural networks processing can be followed. As long as the TI property of wavelet kernels and kernels used in the convolution process is observed, the entire CNN would also possess the same TI property.

2.3.7 Potential application of composing translationally identical and rotationally identical kernels

Considering two type of TI CNN systems: rotationally identical CNN (RTI-CNN) and translationally identical CNN (TLI-CNN) which are mutually exclusive properties. This is to say that a CNN system can only posses either a RTI-CNN or a TLI-CNN, but not both properties in the same CNN. In fact, all different types of TLI-CNN systems are mutually exclusive, one CNN can only possess at most one type of TLI property. Only in some situations, an RTI can be composed by other RTIs. This is out of the scope of this paper.

As indicated above, value for each node on the first flatten layer would be obtained by the last convolution layer with or without transformationally independent function C[.] inner product with K_f (transformation corresponding weights sharing). This computation would produce identical value on each node (n) (i.e., $s_2[n]$) with any type-1 transformation of the input vector. After this layer, identical signals would be processed by the same network and produce an identical output vector, by entering either the Vi or its T**{** Vi **}**. Using 2D Dih4 TI as an example, T{Vi} has 7 type-1 transformation versions beside the original, if stride and pooling are not used. For a 2D CNN, translation of the input vector along 45° line, there are as many versions as wish. Again, there are computationally symmetric inside the CNN structure and would result in identical values by arranging share weights at each node on the first flatten layer. Hence there is no need to recompute providing all kernels possess the same TI property.

Considering a 2D CNN system, it consists of 5 groups of convolution processes: 1 process with Dih4 RTI kernels and 4 processes with 4 different types of TLI kernels as shown in Figure 2. Each of these 5 processes connects to a different group of flatten nodes. In other words, $s_2(n)$ discussed above would become $s_2[m, n_m]$, where m = 1,2, .. 5 group and each flatten node group $n_m$ = 1,2,...$N_m$, and $N_m$ is the total number of nodes in group m. These 5 groups of flatten nodes can be fully connected to the following flatten layer or the output layer which is fully connected without sharing weights.

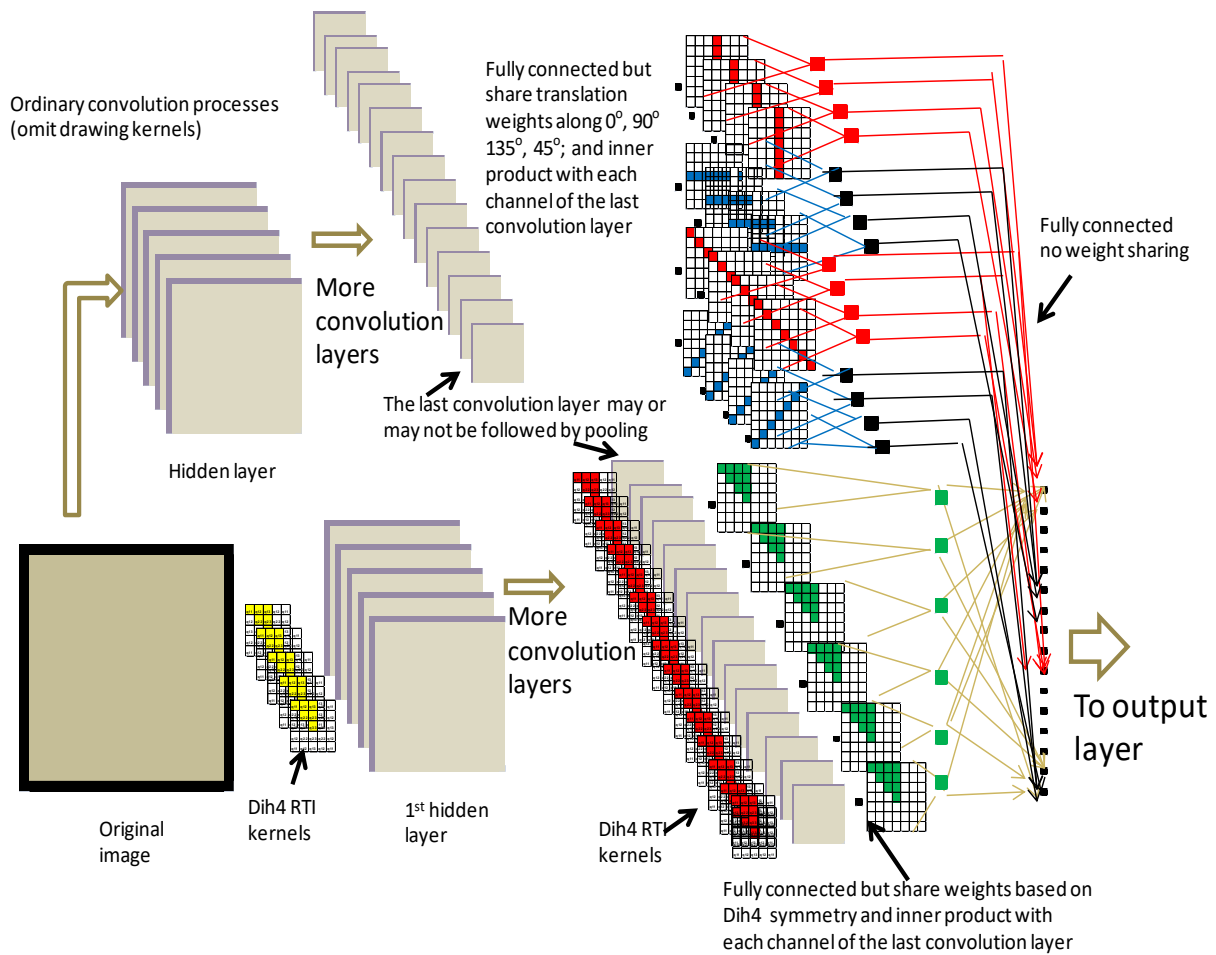

Figure 2: A CNN contains 1 Dih4 TI pipeline and 4 TLI pipelines. Each of them carries its corresponding TI property up to its first flatten layer. Each node of the corresponding channel banks serves as a feature extraction terminal before entering into the classification section of the CNN.

Because different TI properties are mutually exclusive literally, a node in the layer that fully connected to 5 (or any two) groups with different TI convolution processes does not possess TI any more. But each of them would possess specific feature extraction for the construction of a more coordinated CNN system.

The number of free parameters in these kernels are 3-6 times less than its ordinary CNN kernel counterparts. Each group of TI kernels should follow its TI pipeline until the last convolution layer and would still possess the same TI property. Each TI pipeline would be merged into one of 5 independent channel banks on the first flatten layer **that serves as feature extraction terminal before entering classification part of the CNN**. Several ways to train this type of composed CNN.

(a) Train one pipeline at a time and merge their feature banks in the second run of training.

(b) Train several pipelines together and merge all feature banks in the second run of training.

(c) Train all pipeline together.

It will take some time for investigators in various fields to find out the performance of this type of TI-CNN systems or their composition. The CNN with composed TI kernels would possess several TI properties on the channel banks of the first flatten layer. However, the output layer would not receive identical signals while entering a type-2 transformation version of the input vector discussed earlier. However, the advantage of using the composed CNN pipelines, each with a type of TI kernels in a CNN, is that it would be trained by both multiple directional kernels and Dih4 (wedge shape symmetry) kernels and is expected to produce more stable result by entering an input vector (Vi) and its rotation versions as well as its translation versions. It is conceivable that these TI kernels would assist investigators in finding what are the filters the CNN has learned after a course of CNN training and may help explain reasons of its learning processes instead of taking them as black boxes.

## 2.4. Samples of Computer Implementation for Major Components of TI-CNN

Computer C/C++ codes for the implementation of RTI-CNN and TLI-CNN have been made by the authors. Because many types of TI-CNN can be formed, only most desired few samples are provided in the appendix section of this paper for investigators' convenience and facilitate the applications of the TI-CNN. Because the TI-CNN requires a symmetric platform, the shape of vector and kernel should be a square in 2D and a cubic in 3D as demonstrated in the computer programming implementation (see Appendices A and B).

### 2.4.1. 2D RTI-CNN

By assigning the initial pointer of a TI kernel with size of (ksz_1d × ksz_1d) to the computer memory address (float *kernel), multiply the kernel with a kernel size area of a 2D vector whose original size is (vsz_1d × vsz_1d); where ksz_1d denotes 1D kernel size and vsz_1d denotes the original 1D vector size. The starting element of the vector in a computer memory address for multiplication is (float *vector). Two C/C++ routines (one for Dih4 symmetry and the other is for $90^o$ rotation symmetry) are provided in Appendix A. Each routine returns the summation of products from corresponding element multiplications. The user should make sure that the computer memory pointer to the first element of vector is within the vector matrix of (vsz_1d - ksz_1d +1 ) × (vsz_1d - ksz_1d + 1) which would also be the resultant matrix size.

Usually the vector size is greater than the kernel size (i.e., vsz_1d > ksz_1d). The fan-in connection element on the resultant layer receives the summation of products. Then the same computation process takes place to the next stop of the scan. However, the vector size is equal to the kernel size (i.e., ksz_1d = vsz_1d) as an inner product calculation and each fan-in connection node on the fatten layer (sharing weights not as fully connection weights in an ordinary CNN) receives the summation of all multiplication products.

In the error backpropagation process, each kernel weight would be altered accordingly based on its element position in the kernel. For a TI kernel, a large part of coefficients share the same weight with their transformation corresponding elements. Since each element in the kernel is updated by adding the propagated error modification, all their transformation corresponding element counterparts should also receive others' changes, so that the updated kernel would maintain TI property. However, averaging values contributed from all corresponding symmetric elements can be a better choice in some practices. Right after this process, each element of the kernel should be updated as: *(kernel + v*msz_1d + u) += *(matrixP + v*msz_1d + u); in C programming language.

Two C/C++ routines that perform the coefficient changes for each element in the kernel is provided in Appendix B. Two modes are available for using these routines: summing up and averaging. For initialization of weights in a TI kernel, averaging mode or directly assign the same value on the corresponding symmetric positions would be a rational choice.

2.4.2. Instruction for Conversion of an ordinary 2D CNN to 2D TI-CNN

There are 3-5 places in the CNN program that should be altered in order to convert an ordinary 2D CNN to a 2D TI-CNN depending on the computer program architecture of the CNN. Let use a Dih4 TI system as an example.

(a) Initialization of convolution weights -

After each kernel given a set of initial weights, the program should be inserted a statement

   turnMx_to_Dih4RTI_2D(0, ksz_1d, kernel);

so that each kernel vector would possess the Dih4 TI property.

(b) In the forward propagation of a CNN: All inner product of a convolution step can be replaced by the following computer statement

   sum_products = vectorTkernel_Dih4RTI_2D(vsz_1d, vector, ksz_1d, kernel);

Since the kernel possesses a TI property (i.e., $K = T\{K\}$), the above computer statement is not necessary to replace the inner product of an ordinary convolution process. The difference between them is their computational effectiveness. There is no difference in their computation results.

For each node on the first flatten layer which fully connects to its operand vector through each connection weight (a 2D kernel) before the first flatten layer, the 2D vector size (vsz_1d x vsz_1d) is equal to the 2D kernel size (ksz_1d x ksz_1d). The kernel, possesses fully connected weights, would performs an inner product with the vector using the same computer statement shown above. Unlike the process in the convolution layer, this process must be taken place to ensure the quantitatively identical result be observed at each node of the first flatten layer, with respect to all associated transformations.

(c) In the error-backpropagation process of a CNN: After modification of weights for all kernel elements is completed and collected in a temporary memory kernel_C, enter the following computer statements

   turnMx_to_Dih4RTI_2D(1, ksz_1d, kernel_C);

```
    for (v = 0; v < ksz_1d; v++) {
       for (u = 0; u < ksz_1d; u++) {
         *(kernel + v*ksz_1d + u) += *(kernel_C + v*ksz_1d + u);
       }
    }
```

For each node on the first flatten layer, again the 2D vector size (vsz_1d x vsz_1d) is equal to the 2D kernel size (ksz_1d x ksz_1d) that is also the total number of fully connected weights. Hence there are vsz_1d x vsz_1d number of fan-in weights. After the modification of weights for all kernel elements is completed and collected in a temporary memory kernel_C, enter the above computer statements.

Comments:
(A) With a TI kernel property throughout the entire TI-CNN, the total number of effective parameters on each kernel would be reduced by approximately the factor of the TI symmetry order.  Depending upon application, the TI-CNN may need to increase the same factor in number of channels to compensate the less number of parameters in the kernel. This would maintain the total number of nonlinear equations in the neural networks to be solved by the training.

(B) If summing up changes of error-backpropagation is used for all transformation sharing elements, the training may or may not jump to an uncontrollable situation. In such a case, an adjustment of gain in the error-backpropagation may be useful in some applications.

(C) It is not suitable to use a TI-CNN for an application that requires differentiation orientation patterns of the input vector. Because it conflicts with the intended TI property in a TI-CNN.

 (D) In the above instruction, routines "vectorTkernel_Dih4RTI_2D(...);" and "turnMx_to_Dih4RTI_2D(...);"  can be replaced by "vectorTkernel_90dgRTI_2D(...) and "turnMx_to_90dgRTI_2D(...);" for forward propagation and backpropagation, respectively. The CNN would then be converted to a 90 degree rotation TI CNN system.

**C. Experiments and Initial Results of 2D TI-CNNs**
1. Random Number Generator Made Vectors as Input Processed with Random number RTI Kernels in a Typical CNN
      To confirm our approach and to validate C/C++ computer programming routines for various CNN structures composing of convolution kernels with and without involving stride, pooling, 1D convolution across the channels in a CNN layer, we randomly generated 2D matrix Vi(x,y) and 2D kernels $K_{l,c}(u,v)$ with

$$T_{tr}\{ \ K_{l,c}(u,v) \ \} = K_{l,c}(u,v) \qquad \qquad ...(24)$$

where "l" denotes layer number and "c" denotes channel number in layer "l", $T_{tr}\{\}$ transformation of a matrix with either $90°$, $180°$, $270°$, or $360°$ rotation and their corresponding reflection counterparts. The size of Vi ranged from 10 to 1000 in 1D and kernel size ranged from 3 to 30 in 1D, number of convolution layers ranged from 1 to 10. The size of channel matrices in any hidden convolution layer are evenly divisible by the stride length and its immediate pooling size.
      The result indicated that the difference between two output vectors: CNN[ Vi ] - CNN[$T_{tr}\{$ Vi $\}$] vanished (i.e., $\vec{0}$ ) for all tests with type-1 rotation and their reflection and all sizes of $K_{l,c}(u,v)$ that contain a set of random numbers but satisfy eq. (24).

We also replaced all kernels with Dih4 symmetry and tested on a small lung nodule data set. We found that during the training and after the training, the result of CNN[ Vi ] - CNN[T$_{tr}${ Vi }] remains vanished (i.e., $\vec{0}$).

2. 2D CNN processed with stride and the pooling to test translation of the input

When a vertical size of a hidden channel matrix cannot be evenly divided by either the stride vertical length or the pooling vertical size, the CNN maintained TI only with reflection transformation about x=0 axis (i.e., only CNN[ Vi ] - CNN[ T$_{x=0}${ Vi } ] = $\vec{0}$; otherwise the difference would not be vanished). When a horizontal size of a hidden channel matrix cannot be evenly divided by either the stride horizontal length or the pooling horizontal size, the CNN maintained TI only with reflection transformation about y=0 axis (i.e., only CNN[ Vi ] - CNN[T$_{y=0}${ Vi }] = $\vec{0}$; otherwise the difference would not be vanished). A simple remedy to this issue is to zero pad the intermediate resultant matrices for processing each operation symmetrically in the operating space.

**D. Conclusions and Discussion -**

An ordinary CNN using unconstraint kernel weighs would fully depend on the backpropagation training. The use of rotation and translation versions of each input vector as data argumentation has been proposed since the CNN technique was independently developed by the authors [Lo 1993]. However, many investigators reported that the method has limited success with a long training time. The proposed TI-CNN technique is intended to improve the CNN capability in terms of generalization and reduction of the training time.

One particular networking process must be taken care to maintain the same TI property is that signals from the channel of last convolution layer fan-in connecting to each node of the first flatten layer. This can be done by treating the flatten layer as another convolution layer. Each node of the flatten layer is taken as a channel in the forward propagation. Each channel at the flatten layer would perform inner product with its TI kernel and channels of the previous layer and produce a resultant matrix (n×n×n element; where n =1). This would possess the same transformationally symmetric property for the first flatten layer. Since transformation of a single element functions as an identity, the value receives in each node of the first flatten layer would be the same, so as all network nodes in the following pipeline, with respect to all corresponding transformations of the original input vector. Therefore, the CNN possessing the same TI property among all convolution layers and the first flatten layer would output the same result for all transformation versions of the original input vector. One issue that can cause loss of TLI property in a CNN is the stride and/or pooling in translation. Only those translation distances that are evenly divisible by the effective length of stride distance times pooling size, can maintain the translationally identical property in the CNN.

Lo and his co-workers (Lo is the first author of this paper) had employed 1D TI property on multiple circular path CNN for the classification of breast lesions and lung nodules (Lo 1998; 2002). It was a long overdue for the authors to elaborate full TI properties in 2D and 3D geometries, though the fundamental theory of TI and its process within the CNN has been realized by the authors during the initial development of the circular path CNN (~1997).

With or without using a family of TI kernels, type-2 transformation of input vector is still needed as a part of data augmentation for the intention of training the CNN to produce qualitatively invariant output consistently, if nearly identical result is not obtained. For the use of each type-2 transformation, it automatically generates corresponding number of transformation versions for training the TI-CNN. The total number of data augmentation can be significantly reduced as compared to that of the conventional CNN. The most usefulness of the proposed transformationally identical technique is that one can insert a few key computer routines in an ordinary or a custom-designed CNN so that the CNN would automatically turn itself into a TI-CNN. Translation and rotation TI kernels can also be composed together to train the CNN. In order to facilitate the use of TI-CNN, two sets of C programming language routines are provided in the Appendices to facilitate AI investigators' evaluation of the proposed TI-CNN techniques in diversified applications when a TI property is desired.

**Appendix A:**
A.1. A C programming language routine that performs a 2D Dih4 symmetric kernel inner product with a corresponding region on a 2D vector

```
/* ************************************************************************ */
/* C Routine - a 2D Dih4 symmetric kernel inner product with a region on a 2D vector */
/* 2D vector entry address: (vector);    2D kernel initial element address: (kernel)    */
/* 1D vector size: vsz_1d;                1D kernel size: ksz_1d                         */
/* Author: ShihChung Benedict Lo, Ph.D.                                                  */
/* Creation date: 4/30/2018                                                              */
/* Version 1    (© , 2018, ShihChung Benedict Lo – All Rights Reserved )                 */
/* ************************************************************************ */
float vectorTkernel_Dih4RTI_2D(int vsz_1d, float *vector, int ksz_1d, float *kernel)
{
  int y, x, yy, xx, odd_even, ksz_1d_d2;
  int k_yr, yr, yyr, yrx, yyrx, yrxx, yyrxx, xr, xxr, xry, xryy, xxry, xxryy;
  float sum_products, tmp;

  odd_even = ksz_1d%2;
  ksz_1d_d2 = (ksz_1d-1) / 2;
  sum_products = 0.;

  k_yr= 0;  yr = 0;  yyr = vsz_1d * (ksz_1d-1);
  for (y = 0, yy = ksz_1d - 1; y <= ksz_1d_d2; y++, yy--) {
    xr = yr;  xxr = yyr;
    for (x = y, xx = ksz_1d - 1 - y; x <= ksz_1d_d2; x++, xx--) {
      yrx = yr + x;
      if (y == x) {
        if (y != ksz_1d_d2 || odd_even != 1) {
          yrxx = yr + xx;   yyrx = yyr + x;   yyrxx = yyr + xx;
          tmp = ( *(vector + yrx) + *(vector + yrxx) + *(vector + yyrx) + *(vector + yyrxx) );
        } else {
          tmp = *(vector + yrx);
        }
      } else if (x == ksz_1d_d2 && odd_even == 1) {
        if (y != ksz_1d_d2) {
          xry = xr + y;   xryy = xr + yy;   yyrx = yyr + x;
          tmp = ( *(vector + yrx) + *(vector + xry) + *(vector + yyrx) + *(vector + xryy) );
        } else {
          tmp = *(vector + yrx);
        }
      } else {
        yrxx = yr + xx;   yyrx = yyr + x;   yyrxx = yyr + xx;
        xry = xr + y;   xryy = xr + yy;   xxry = xxr + y;   xxryy = xxr + yy;
        tmp = (*(vector + yrx) + *(vector + yrxx) + *(vector + yyrx) + *(vector + yyrxx)
             + *(vector + xry) + *(vector + xryy) + *(vector + xxry) + *(vector + xxryy));
      }
      sum_products += (tmp * *( kernel + k_yr + x));
      xr += vsz_1d; xxr -= vsz_1d;
    }
    k_yr += ksz_1d;
```

```c
    yr += vsz_1d; yyr -= vsz_1d;
  }
  return (sum_products);
}
```

## A.2.  A C programming language routine that performs a 2D 90-degree rotationally symmetric kernel inner product with a corresponding region on a 2D vector

```c
/* ***************************************************************************** */
/* C Routine - a 2D 90-degree RI kernel inner product with a region on a 2D vector   */
/* 2D vector entry address: (vector);    2D kernel initial element address: (kernel)    */
/* 1D vector size: vsz_1d;               1D kernel size: ksz_1d                      */
/* Author: ShihChung Benedict Lo, Ph.D.                                              */
/* Creation date: 4/30/2018                                                          */
/* Version 1    (© , 2018, ShihChung Benedict Lo – All Rights Reserved )            */
/* ***************************************************************************** */
 float vectorTkernel_90dgRTI_2D(int vsz_1d, float *vector, int ksz_1d, float *kernel)
{
  int y, x, yy, xx, odd_even, ksz_1d_d2;
  int k_yr, yr, yrx, xr, xry, xryy, yyr, yyrxx, xxr, xxry, xxryy, tyyr;
  float sum_products, tmp;

  odd_even = ksz_1d%2;
  k_yr = 0; ksz_1d_d2 = (ksz_1d-1) / 2;
  sum_products = 0.;

  yr = 0; tyyr = yyr = vsz_1d * (ksz_1d-1);
  for (y = 0, yy = ksz_1d - 1; y <= ksz_1d_d2; y++, yy--) {
    xr = 0; xxr = tyyr;
    for (x = 0, xx = ksz_1d - 1; x <= ksz_1d_d2; x++, xx--) {
      if(x == ksz_1d_d2 && odd_even == 1) {
        if (y != ksz_1d_d2) {
          yrx = yr + x;  xry = xr + y;  yyrxx = yyr + xx;  xxryy = xxr + yy;
          tmp = ( *(vector + yrx) + *(vector + xry) + *(vector + yyrxx) + *(vector + xxryy) );
          sum_products += (tmp * *(kernel + k_yr + x));
        } else {
          sum_products += (*(vector + yr + ksz_1d_d2) * *(kernel + k_yr + ksz_1d_d2));
        }
      } else {
        yrx = yr + x;  xryy = xr + yy;  yyrxx = yyr + xx;  xxry = xxr + y;
        tmp = ( *(vector + yrx) + *(vector + xryy) + *(vector + yyrxx) + *(vector + xxry) );
        sum_products += (tmp * *(kernel + k_yr + x));
      }
      xr += vsz_1d; xxr -= vsz_1d;
    }
    k_yr += ksz_1d;
    yr += vsz_1d; yyr -= vsz_1d;
  }
  return (sum_products);
}
```

## Appendix B:
B.1.  A C programming language routine that turns a 2D matrix to a 2D Dih4 symmetric matrix.

```c
/* ************************************************************************************* */
/* C Routine - Turn a 2D matrix to a 2D Dih4 symmetric matrix                            */
/* 2D matrix initial element address: (matrixP)                                          */
/* 1D matrix size: msz_1d                                                                */
/* "mode == 1" sum up values and return to all corresponding elements; "else" average.   */
/* Author: ShihChung Benedict Lo, Ph.D.                                                  */
/* Creation date: 4/30/2018                                                              */
/* Version 1     (© , 2018, ShihChung Benedict Lo – All Rights Reserved )                */
/* ************************************************************************************* */
void  turnMx_to_Dih4RTI_2D (int mode, int msz_1d, float *matrixP)
{
  int v, u, vv, uu, odd_even, msz_1d_d2, dv1, dv2;
  int vr, vvr, vru, vvru, vruu, vvruu, ur, uur, urv, urvv, uurv, uurvv;
  float sum, *C;

  C = matrixP;
  if (mode == 1) { /* This mode would perform summing up all change values */
    dv2 = dv1 = 1;
  } else { /* Else, it would take average of change values corresponding to element positions */
    dv2 = 8;
    dv1 = 4;
  }
  odd_even = msz_1d%2;
  msz_1d_d2 = (msz_1d-1) / 2;

  vr = 0;  vvr = msz_1d * (msz_1d-1);
  for (v = 0, vv = msz_1d - 1; v <= msz_1d_d2; v++, vv--) {
    ur = vr;   uur = vvr;
    for (u = v, uu = msz_1d - 1 - v; u <= msz_1d_d2; u++, uu--) {
      vru = vr + u;
      if (v == u) {
        if (v != msz_1d_d2 || odd_even != 1) {
          vruu = vr + uu;  vvru = vvr + u;   vvruu = vvr + uu;
          sum = ( *(C + vru) + *(C + vruu) + *(C + vvru) + *(C + vvruu) );
          *(C + vru) = *(C + vruu) = *(C + vvru) = *(C + vvruu) =  sum / dv1;
        }
      } else if (u == msz_1d_d2 && odd_even == 1) {
        if (v != msz_1d_d2) {
          urv = ur + v;  urvv = ur + vv;  vvru = vvr + u;
          sum = ( *(C + vru) + *(C + urv) + *(C + vvru) + *(C + urvv) );
          *(C + vru) = *(C + urv) = *(C + vvru) = *(C + urvv) = sum / dv1;
        } /* else { Center of odd size } */
      } else {
        vruu = vr + uu;   vvru = vvr + u;   vvruu = vvr + uu;
        urv = ur + v;   urvv = ur + vv;   uurv = uur + v;   uurvv = uur + vv;
        sum = ( *(C + vru) + *(C + vruu) + *(C + vvru) + *(C + vvruu)
              + *(C + urv) + *(C + urvv) + *(C + uurv) + *(C + uurvv) );
        *(C + vru) = *(C + vruu) = *(C + vvru) = *(C + vvruu)  =
```

```
       *(C + urv) = *(C + urvv) = *(C + uurv) = *(C + uurvv)  = sum / dv2;
      }
      ur += msz_1d; uur -= msz_1d;
    }
    vr += msz_1d; vvr -= msz_1d;
  }
}
```

## B.2.  A C programming language routine that turn a 2D matrix to a 2D 90-degree rotationally identical matrix .

```
/* ************************************************************************************ */
/* C Routine - Turn a 2D matrix to a 2D 90-degree rotationally identical matrix     */
/* 2D matrix initial element address: (matrixP)                                     */
/* 1D matrix size: msz_1d                                                            */
/* "mode == 1" sum up values and return to all corresponding elements; "else" average */
/* Author: ShihChung Benedict Lo, Ph.D.                                              */
/* Creation date: 4/30/2018                                                          */
/* Version 1     (© , 2018, ShihChung Benedict Lo – All Rights Reserved )            */
/* ************************************************************************************ */
void  turnMx_to_90dgRTI_2D(int mode, int msz_1d, float *matrixP)
{
  int v, u, vv, uu, odd_even, msz_1d_d2, dv;
  int  vr, vru, ur, urv, urvv, vvr, vvruu, uur, uurv, uurvv, tvvr;
  float sum,  *C;

  C = matrixP;
  dv = (mode == 1) ? 1 : 4;
  odd_even = msz_1d%2;
  msz_1d_d2 = (msz_1d-1) / 2;

  vr = 0; tvvr = vvr = msz_1d * (msz_1d-1);
  for (v = 0, vv = msz_1d - 1; v <= msz_1d_d2; v++, vv--) {
    ur = 0; uur = tvvr;
    for (u = 0, uu = msz_1d - 1; u <= msz_1d_d2; u++, uu--) {
      if( u == msz_1d_d2 && odd_even == 1) {
        if(v != msz_1d_d2) {
           vru = vr + u;  urv = ur + v;  vvruu = vvr + uu;  uurvv = uur + vv;
           sum = ( *(C + vru) + *(C + urv) + *(C + vvruu) + *(C + uurvv) );
           *(C + vru) = *(C + urv) = *(C + vvruu) = *(C + uurvv) = sum / dv;
        } /* else { Center of odd size } */
      } else {
         vru = vr + u;  urvv = ur + vv;  vvruu = vvr + uu;  uurv = uur + v;
         sum = ( *(C + vru) + *(C + urvv) + *(C + vvruu) + *(C + uurv) );
         *(C + vru) = *(C + urvv) = *(C + vvruu) = *(C + uurv) = sum / dv;
      }
      ur += msz_1d; uur -= msz_1d;
    }
    vr += msz_1d; vvr -= msz_1d;
  }
}
```